# A Novel lightweight Convolutional Neural Network – ExquisiteNetV2

JOU, Shyh-Yaw,     SU, Chung-Yen

**Abstract**

In the paper of ExquisiteNetV1, the ability of classification of ExquisiteNetV1 is worse than DenseNet.

In this article, we propose a faster and better model ExquisiteNetV2. We conduct many experiments to evaluate its performance. We test ExquisiteNetV2, ExquisiteNetV1 and other 9 well-known models on 16 credible datasets including Cifar-10 under the same condition. According to the experimental results, ExquisiteNetV2 gets the highest classification accuracy over half of the datasets. Important of all, ExquisiteNetV2 has fewest amounts of parameters. Besides, in most instances, ExquisiteNetV2 has fastest computing speed.

**Model Architecture**

ExquisiteNetV2 mainly consists of 4 types of blocks, namely DFSEBV2 block (DFSEB[1] version 2), ME block[1], FCT block and EVE block. Besides, we design a new kind of SE block[2], that is, SE-LN block.

*A. EVE Block*

Complete name is extreme-value-expansion block.

In the research of ExquisiteNetV1[1], we found using max pooling will get better accuracy than using depthwise convolution with its stride set to 2. We are very curious about why max pooling also work well despite it discard the 75% features when downsampling.

According to Fig 1, after applying max/min pooling to a cute Corgi image, we can find that the result image of max pooling look brighter and the one of min pooling look darker. It is reasonable because max/min pooling will only retain big/small pixel values. Important of all, we can find that despite max/min pooling will discard 75% feature, the both result images still look much like original image, in other words, the feature retained by max/min pooling might still be able to represent the original feature.

To make ME block of ExquisiteNetV1 be able to retain more feature when downsampling, we combine a min pooling layer into ME block. So the amount of discard feature down to 50%. Because revised ME block retains the extreme feature, we rename it as EVE block. The architecture of EVE block is shown in Fig 2.

We also make a try to combine average pooling layer into EVE block to retain more feature. However, after adding the average pooling layer into EVE block, the accuracy become worse.

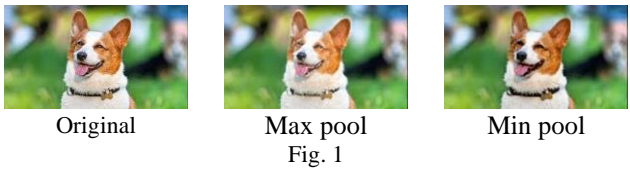

Original    Max pool    Min pool

Fig. 1

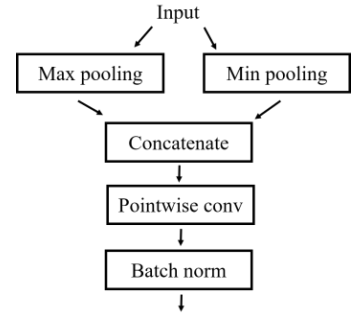

Fig. 2 EVE block

*B. FCT Block*

Complete name is Feature-Concentrator block.

The feature in the raw image is the most important so we should try to retain feature of raw image as many as possible.

In order to retain all feature of raw image, we combine the EVE block with a 4×4 depthwise conv and use FCT as input block of ExquisiteNetV2. The architecture of FCT block is shown in Fig 3.

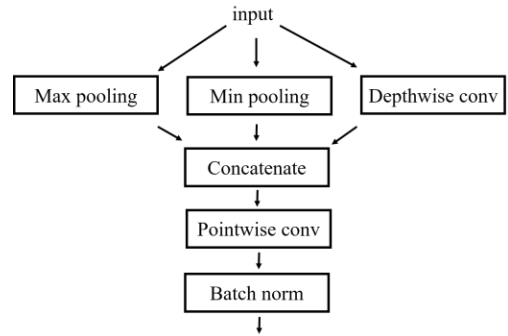

Fig. 3  FCT block

*C. SE-LN Block*

Complete name is SE-Layer-Normalization block.

SE block include 2 fully-connected layer, we use layer normalization to replace these 2 fully-connected layer for decrease amount of parameters. The amounts of parameters of layer normalization is much fewer than ones of 2 fully-connected layer[3]. The architecture of SE-LN block is shown in Fig 4.

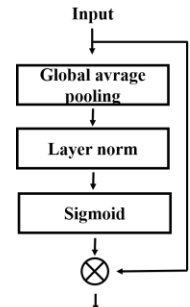

Fig. 4 SE-LN block

*D. DFSEBV2 Block*

We delete one SE block and replace the other SE blcok into SE-LN block so the amounts of parameters is fewer than DFSEBV1. Besides, we change some activation. The architecture of DFSEBV2 block is shown in Fig 5.

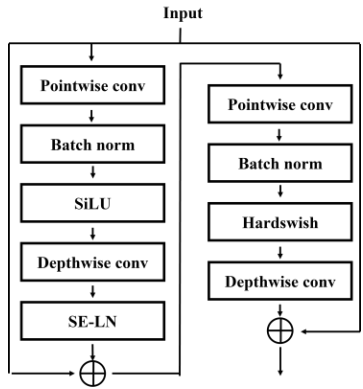

Fig. 5 DFSEBV2 block

*E. ExquisiteNetV2*

We show the architecture in Table. 1.

We only replace the first and second ME block in ExquisiteNetV1 with FCT and EVE block. One reason is that the amounts of parameters in ME block is fewer than FCT or EVE block, the other reason is that we think the raw feature is the most important so we only use FCT or EVE block in the front part of model. Besides, we change the pointwise conv in ExquisiteNetV1 into depthwise conv. The amounts of parameters of ExquisiteNetV2 is fewer than ExquisiteNetV1. The parameters of each model is shown on Table. .

Table. 1 Architecture of ExquisiteNetV2

| Input | Operator | Out Channels |
|---|---|---|
| $224^2 \times 3$ | FCT | 12 |
| $112^2 \times 12$ | DFSEBV2 | 12 |
| $112^2 \times 12$ | EVE | 48 |
| $56^2 \times 48$ | DFSEBV2 | 48 |
| $56^2 \times 48$ | ME | 96 |
| $28^2 \times 96$ | DFSEBV2 | 96 |
| $28^2 \times 96$ | ME | 192 |
| $14^2 \times 192$ | DFSEBV2 | 192 |
| $14^2 \times 192$ | ME | 384 |
| $7^2 \times 384$ | DFSEBV2 | 384 |
| $7^2 \times 384$ | Depthwise Conv | 384 |
| $7^2 \times 384$ | Hard Swish | 384 |
| $7^2 \times 384$ | Average pooling | 384 |
| $1^2 \times 384$ | Dropout | 384 |
| $1^2 \times 384$ | FC | - |

**Dataset**

We choose 16 credible datasets in the experiment, namely DTD, Chest X-Ray, OCT, Leukemia, RAF-DB, SUN397, Fruits-360, Food-101, Stanford Online Products, Caltech256, STL-10, Birds-2011, VGG Flowers, Plant SeedlingsV2, Synthetic Digits and Cifar-10 [4]-[18].

1. DTD

It is a texture dataset. The resolution of every image range from 300×300~640×640.

Table. 2 the information of DTD

|  | Original | Use in experiment |
|---|---|---|
| Training set | 1880 | |
| Test set | 1880 | |
| Classes | 47 | |

2. Chest X-Ray

This dataset is used to recognize whether the children have the lung disease.

Table. 3 the information of Chest X-Ray

|  | Original | Use in experiment |
|---|---|---|
| Training set | 5232 | |
| Test set | 624 | |
| Classes | 2 | |

3. OCT

OCT is Optical Coherence Tomograph. The classes are diabetic macular edema, choroidal neovascularization, drusen and normal. The size of training set is so big that we randomly sample 250 images from every class.

Table. 4 the information of OCT

|  | Original | Use in experiment |
|---|---|---|
| Training set | 207130 | 1000 |
| Test set | 1000 | |
| Classes | 4 | |

4. Leukemia

This dataset is used to recognize whether the children have Leukemia.

Table. 5 the information of Leukemia

|  | Original | Use in experiment |
|---|---|---|
| Training set | 10661 | |
| Test set | 1867 | |
| Classes | 2 | |

5. RAF-DB

Every image is human's facial expression. There are 2 sets in the original dataset, one is basic, the other is compound. We train and test on basic.

Table. 6 the information of RAF-DB

|  | Original | Use in experiment |
|---|---|---|
| Training set | 12271 | |
| Test set | 3068 | |
| Classes | 7 | |

6. SUN397

The dataset contains various scene images.

Table. 7 the information of SUN397

|  | Original | Use in experiment |
|---|---|---|
| Training set | 19850 | |
| Test set | 19850 | |
| Classes | 397 | |

7. Fruits-360

The size of training set is so big that we randomly sample 30 images from every class.

Table. 8 the information of Fruits-360

|  | Original | Use in experiment |
|---|---|---|
| Training set | 67692 | 3930 |
| Test set | 22688 | |
| Classes | 131 | |

8. Food-101

The size of training set is so big that we randomly sample 50 images from every class.

Table. 9 the information of Food-101

|  | Original | Use in experiment |
|---|---|---|
| Training set | 75750 | 5050 |
| Test set | 25250 | |
| Classes | 101 | |

9. Stanford Online Products

The size of training set is so big that we randomly sample 300 images from every class.

Table. 10 the information of Stanford Online Products

|  | Original | Use in experiment |
|---|---|---|
| Training set | 59551 | 3600 |
| Test set | 60502 | |
| Classes | 12 | |

10. Caltech256

The name is "256", but there is actually 257 classes in the dataset. The 257th class is "clutter". We think the 257th class is so unclear that we delete this class. The original dataset is not divide into training set and test set, so we divide each class into 10 equal parts. One of the parts is used as training set, the rest are test set.

Table. 11 the information of Caltech256

|  | Original | Use in experiment |
|---|---|---|
| Training set | 30607 | 2881 |
| Test set |  | 26899 |
| Classes | 257 | 256 |

11. STL-10

The dataset is inspired from CIFAR-10[19]. Because the resolution of the image in CIFAR-10 is too low(32×32) [19], the images in STL-10 are all 96×96.

Table. 12 the information of STL-10

|  | Original | Use in experiment |
|---|---|---|
| Training set | 5000 | |
| Test set | 8000 | |
| Classes | 10 | |

12. Birds-2011

The dataset includes various bird species.

Table. 13 the information of Birds-2011

|  | Original | Use in experiment |
|---|---|---|
| Training set | 5794 | |
| Test set | 5994 | |
| Classes | 200 | |

13. VGG Flowers

The dataset includes various flower species.

Table. 14 the information of VGG Flowers

|  | Original | Use in experiment |
|---|---|---|
| Training set | 1020 | |
| Test set | 6149 | |
| Classes | 102 | |

14. Plant SeedlingsV2

The dataset contains 12 kinds of crop seed. Because the original dataset is not divide into training set and test set, so we divide each class into 2 equal parts. One is used as training set, the other is test set.

Table. 15 the information of Plant SeedlingsV2

|  | Original | Use in experiment |
|---|---|---|
| Training set | 5539 | 2766 |
| Test set |  | 2773 |
| Classes | 12 | |

15. Synthetic Digits

Every image in this dataset is RGB and their background is complicated.

Table. 16 the information of Synthetic Digits

|  | Original | Use in experiment |
|---|---|---|
| Training set | 10000 | |
| Test set | 2000 | |
| Classes | 10 | |

16. Cifar-10

Every image is 32x32. Many paper use this dataset as benchmark. We randomly split 10000 image from training set as validation set.

Table. 17 the information of Synthetic Digits

|  | Original | Use in experiment |
|---|---|---|
| Training set | 50000 | |
| Test set | 10000 | |
| Classes | 10 | |

**Experiment and Analysis**

*A. Settings*

To make the experimental result being objective, we test each model under the same conditions. The conditions are shown below.

1. Program is run on Ubuntu 16.04
2. Accelerator is GeForce RTX 2080 Ti
3. Every image is converted to RGB
4. Every image is resized to 224×224
5. Every pixel value in image is scaled into [0,1]
6. Each dataset is not augmented (only cifar-10 augment)
7. Every model is trained without pretrained weight, each weight in each layer is initialized by Pytorch default method.
8. Every model is test with 2 different optimizer respectively. One is RangerVA[21], the other is SGD[22].
9. We have set random seed so the images are read in the same order by each model
10. Batch size is 50 (cifar-10 is 64)
11. Initial learning rate is 0.05 and the factor is 0.1. If the training loss is not improved for consecutive 5 epochs, the

learning rate will be multiplied by factor. Training loss is the average loss of each image in the training set.
12. To avoid the overfitting phenomenon, we add the Dropout layer before the last fully-connected layer into each model.
13. End condition is training loss < 0.01 or total epochs reach 150. The reason why we design this way is that it is difficult to make training loss lower than 0.01 for some models.
14. We have test whether the SE-LN block is able to replace the SE block. We set the decrease ratio of each SE block to 3

*B. Classification Results*

We test many models, namely SE-ResNet18[2], MobileNetV3-Large, EfficientNet-b0, DenseNet121, ResNet18, ResNet50, ShuffleNetV2, GhostNet[23]~[28]. "LN-" means the all SE blocks in the original model are replaced by SE-LN blocks.

Table. 18 parameters of each model
(when the last FC layer contains 1000 output channels)

|  | **Parameters (M)** |
|---|---|
| DenseNet121 | 7.9788 |
| EfficientNet-b0 | 5.3 |
| MobileNetV3-L | 5.483 |
| ShuffleNetV2 | 7.3939 |
| ResNet50 | 25.557 |
| ResNet18 | 11.6895 |
| SE-Resnet18 | 12.1515 |
| LN-Resnet18 | 11.6933 |
| GhostNet | 5.1825 |
| SqueezeNet | 1.2354 |
| SE-ExquisiteNetV1 | 1.5618 |
| LN-ExquisiteNetV1 | 1.3329 |
| SE-ExquisiteNetV2 | 1.0285 |
| LN-ExquisiteNetV2 | **0.8993** |

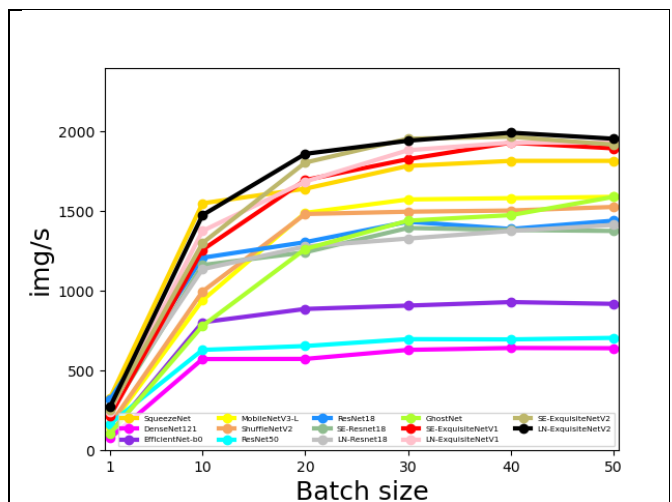

Fig. 6 classifiacation speed of models with different batch size

Table. 19 Test Acc RangerVA(I)

| Optimizer: RangerVA | | | | |
|---|---|---|---|---|
| | Acc (%) | | | |
| | DTD | Chest | OCT | Leukemia |
| DenseNet121 | 27.71 | 90.06 | **91.50** | 75.68 |
| EfficientNet-b0 | 16.97 | 91.51 | 87.60 | 72.52 |
| MobileNetV3-L | 15.21 | 91.83 | 87.60 | 74.77 |
| ShuffleNetV2 | 14.79 | 89.42 | 81.20 | 72.79 |
| ResNet50 | 19.31 | 88.62 | 89.80 | 70.59 |
| ResNet18 | 24.41 | 90.38 | 88.50 | 71.93 |
| SE-Resnet18 | 27.55 | 90.71 | 87.10 | 71.34 |
| LN-Resnet18 | 27.55 | 87.34 | 85.80 | 72.25 |
| GhostNet | 13.72 | 91.35 | 80.00 | 73.54 |
| SE-ExquisiteNetV1 | 32.61 | 89.34 | 83.60 | 76.43 |
| LN-ExquisiteNetV1 | 33.40 | 89.90 | 83.60 | 78.04 |
| SE-ExquisiteNetV2 | **42.66** | **92.15** | 88.40 | 80.66 |
| LN-ExquisiteNetV2 | 40.80 | 91.51 | 90.10 | **82.43** |

Table. 20 Test Acc RangerVA(II)

| Optimizer: RangerVA | | | | |
|---|---|---|---|---|
| | Acc (%) | | | |
| | RAF | SUN397 | Fruit | Food |
| DenseNet121 | **70.57** | **36.89** | 98.26 | 19.14 |
| EfficientNet-b0 | 69.98 | 26.67 | 94.03 | 10.42 |
| MobileNetV3-L | 68.25 | 26.34 | 94.42 | 10.80 |
| ShuffleNetV2 | 66.17 | 26.82 | 93.34 | 10.53 |
| ResNet50 | 65.68 | 28.27 | 96.05 | 12.36 |
| ResNet18 | 68.35 | 28.14 | 96.87 | 14.81 |
| SE-Resnet18 | 68.74 | 28.38 | 98.08 | 18.22 |
| LN-Resnet18 | 67.21 | 27.88 | 97.28 | 15.63 |
| GhostNet | 67.41 | 29.50 | 93.57 | 7.61 |
| SE-ExquisiteNetV1 | 67.99 | 30.04 | 97.95 | 24.60 |
| LN-ExquisiteNetV1 | 65.87 | 28.77 | 97.95 | 24.88 |
| SE-ExquisiteNetV2 | 69.49 | 33.67 | 98.01 | 26.54 |
| LN-ExquisiteNetV2 | 70.08 | 34.44 | 98.10 | **28.07** |

Table. 21 Test Acc RangerVA(III)

| Optimizer: RangerVA | | | | |
|---|---|---|---|---|
| | Acc (%) | | | |
| | Online | Caltech256 | STL10 | Birds |
| DenseNet121 | 38.25 | 20.09 | 72.89 | 38.66 |
| EfficientNet-b0 | 33.62 | 13.65 | 67.33 | 24.19 |
| MobileNetV3-L | 36.12 | 14.44 | 67.30 | 18.27 |
| ShuffleNetV2 | 29.87 | 13.01 | 59.74 | 24.27 |
| ResNet50 | 30.45 | 15.09 | 60.88 | 25.58 |
| ResNet18 | 33.65 | 17.99 | 66.88 | 26.88 |
| SE-Resnet18 | 34.57 | 18.22 | 67.2 | 27.69 |
| LN-Resnet18 | 35.00 | 17.89 | 67.74 | 27.93 |
| GhostNet | 30.10 | 11.22 | 66.84 | 17.98 |
| SE-ExquisiteNetV1 | 37.58 | 20.91 | 70.74 | 35.65 |
| LN-ExquisiteNetV1 | 38.93 | 20.48 | 69.43 | 33.73 |
| SE-ExquisiteNetV2 | 41.58 | 25.06 | 72.24 | 37.04 |
| LN-ExquisiteNetV2 | **43.13** | **25.19** | **73.05** | **39.07** |

Table. 22 Test Acc RangerVA(IV)

| Optimizer: RangerVA | | | |
|---|---|---|---|
| | Acc (%) | | |
| | Flower | Seed | Synthetic Digits |
| DenseNet121 | 40.23 | 94.95 | 98.30 |
| EfficientNet-b0 | 29.74 | 92.32 | 99.50 |
| MobileNetV3-L | 26.26 | 90.23 | 99.55 |
| ShuffleNetV2 | 19.48 | 88.71 | 99.45 |
| ResNet50 | 28.70 | 89.76 | **99.65** |
| ResNet18 | 37.10 | 92.82 | 99.00 |
| SE-Resnet18 | 43.03 | 93.36 | 99.05 |
| LN-Resnet18 | 39.81 | 94.66 | 97.80 |
| GhostNet | 17.52 | 90.59 | 98.45 |
| SE-ExquisiteNetV1 | 50.56 | 95.24 | 98.80 |
| LN-ExquisiteNetV1 | 49.86 | 94.59 | 99.05 |
| SE-ExquisiteNetV2 | 54.95 | 95.56 | 99.50 |
| LN-ExquisiteNetV2 | **55.47** | **95.64** | 99.50 |

Table. 23 Test Acc SGD(I)

| Optimizer: SGD | | | | |
|---|---|---|---|---|
| | Acc (%) | | | |
| | DTD | Chest | OCT | Leukemia |
| DenseNet121 | 26.28 | 87.98 | **90.10** | 72.52 |
| EfficientNet-b0 | 14.31 | 88.62 | 79.70 | 71.02 |
| MobileNetV3-L | 12.93 | 89.26 | 85.80 | 71.67 |
| ShuffleNetV2 | 13.99 | 89.58 | 74.20 | 71.13 |
| ResNet50 | 18.46 | 89.26 | 63.90 | 71.45 |
| ResNet18 | 24.04 | 90.71 | 86.00 | 71.24 |
| SE-Resnet18 | 26.28 | 92.63 | 83.90 | 72.79 |
| LN-Resnet18 | 25.96 | 88.46 | 85.30 | 73.06 |
| GhostNet | 11.33 | 90.22 | 79.70 | 70.86 |
| SE-ExquisiteNetV1 | 33.56 | 90.54 | 78.70 | 75.63 |
| LN-ExquisiteNetV1 | 35.16 | 91.35 | 81.80 | 78.04 |
| SE-ExquisiteNetV2 | 42.23 | 91.83 | 88.80 | 80.61 |
| LN-ExquisiteNetV2 | **42.77** | **93.11** | 88.20 | **83.07** |

Table. 24 Test Acc SGD(II)

| Optimizer: SGD | | | | |
|---|---|---|---|---|
| | Acc (%) | | | |
| | RAF | SUN397 | Fruit | Food |
| DenseNet121 | 68.35 | **37.01** | 97.88 | 20.72 |
| EfficientNet-b0 | 67.89 | 23.63 | 93.73 | 8.55 |
| MobileNetV3-L | 68.32 | 25.07 | 95.83 | 10.28 |
| ShuffleNetV2 | 59.62 | 20.94 | 91.34 | 8.56 |
| ResNet50 | 58.74 | 26.77 | 95.96 | 11.83 |
| ResNet18 | 65.61 | 27.29 | 95.83 | 13.98 |
| SE-Resnet18 | 68.32 | 27.91 | 97.81 | 15.01 |
| LN-Resnet18 | 66.40 | 27.57 | 96.93 | 16.00 |
| GhostNet | 66.33 | 24.69 | 91.73 | 5.97 |
| SE-ExquisiteNetV1 | 64.05 | 32.45 | 98.15 | 26.70 |
| LN-ExquisiteNetV1 | 62.09 | 32.38 | 98.07 | 19.83 |
| SE-ExquisiteNetV2 | **69.20** | 34.97 | 97.88 | 27.21 |
| LN-ExquisiteNetV2 | 69.10 | 34.88 | **98.33** | **28.16** |

Table. 25 Test Acc SGD(III)

| Optimizer: SGD | | | | |
|---|---|---|---|---|
| | Acc (%) | | | |
| | Online | Caltech256 | STL10 | Birds |
| DenseNet121 | 34.02 | 19.10 | 71.31 | 38.54 |
| EfficientNet-b0 | 32.85 | 13.33 | 66.10 | 20.37 |
| MobileNetV3-L | 32.99 | 13.07 | 64.88 | 18.69 |
| ShuffleNetV2 | 27.49 | 11.64 | 57.20 | 16.10 |
| ResNet50 | 27.30 | 14.36 | 57.94 | 25.59 |
| ResNet18 | 34.38 | 17.83 | 66.79 | 26.88 |
| SE-Resnet18 | 34.93 | 19.00 | 65.84 | 30.23 |
| LN-Resnet18 | 32.91 | 17.32 | 67.15 | 27.39 |
| GhostNet | 25.66 | 10.00 | 61.45 | 13.96 |
| SE-ExquisiteNetV1 | 38.20 | 20.12 | 69.46 | 36.22 |
| LN-ExquisiteNetV1 | 38.07 | 20.94 | 68.87 | 33.03 |
| SE-ExquisiteNetV2 | **41.31** | 24.07 | 70.66 | 36.90 |
| LN-ExquisiteNetV2 | 40.68 | **24.89** | **72.20** | **39.26** |

Table. 26 Test Acc SGD(IV)

| Optimizer: SGD | | | |
|---|---|---|---|
| | Acc (%) | | |
| | Flower | Seed | Synthetic Digits |
| DenseNet121 | 38.75 | 95.13 | 98.20 |
| EfficientNet-b0 | 23.45 | 89.51 | 99.65 |
| MobileNetV3-L | 17.79 | 88.28 | **99.80** |
| ShuffleNetV2 | 17.30 | 88.17 | 98.70 |
| ResNet50 | 23.11 | 88.82 | 98.15 |
| ResNet18 | 36.61 | 93.22 | 98.35 |
| SE-Resnet18 | 42.10 | 93.54 | 98.40 |
| LN-Resnet18 | 39.75 | 94.34 | 97.75 |
| GhostNet | 16.96 | 88.21 | 99.30 |
| SE-ExquisiteNetV1 | 50.22 | 95.35 | 98.75 |
| LN-ExquisiteNetV1 | 48.41 | 94.66 | 99.30 |
| SE-ExquisiteNetV2 | 53.88 | 96.21 | 99.05 |
| LN-ExquisiteNetV2 | **56.01** | **96.72** | 99.40 |

Table. 27 the average ranking of each model on 15 datasets

| | RangerVA | SGD |
|---|---|---|
| DenseNet121 | 4.1333 | 4.6666 |
| EfficientNet-b0 | 8.4 | 9.4 |
| MobileNetV3-L | 8.3333 | 8.6666 |
| ShuffleNetV2 | 11.1333 | 11.4666 |
| ResNet50 | 9.3333 | 10.2 |
| ResNet18 | 8.0000 | 7.6666 |
| SE-Resnet18 | 6.8666 | 6 |
| LN-Resnet18 | 8.2666 | 7.4666 |
| GhostNet | 10.6666 | 11 |
| SE-ExquisiteNetV1 | 5.4666 | 4.8 |
| LN-ExquisiteNetV1 | 5.6666 | 4.8 |
| SE-ExquisiteNetV2 | 2.5333 | 2.4666 |
| LN-ExquisiteNetV2 | **1.5333** | **1.4** |

*C. Analysis*

From Table. , we can learn the parameters of LN-ExquisiteNetV2 is fewest. From Fig. , we can learn that as long as the batch size is bigger than 10, the LN-ExquisiteNetV2 is the fastest. The speed is calculated from how long for every model to classify the test set of RAF-DB.

From Table. to Table. , we can learn that SGD is more suitable to ExquisiteNetV2. LN-ExquisiteNetV2 gets first place on 8 datasets when the optimizer is RangerVA and gets first place on 10 datasets when the optimizer is SGD. Besides, LN-ExquisiteNetV2 is much better than state-of-the-art models on some datasets. From Table. , the average ranking of LN-ExquisiteNetV2 is very near 1, it can be seen that LN-ExquisiteNetV2 has strong ability of classification.

However, SE-LN block seems to be only suitable for ExquisiteNet especially ExquisiteNetV2.

*D. Result on Cifar-10*

Table. 28 Cifar-10 result of ExquisiteNetV2
(Train without pretrained weight)

| Val Acc(%) | 93.29 |
|---|---|
| Test Acc(%) | 92.52 |

## E. Object Detection

YoloV5 is one of good object detection models[29]. YoloV5 contains 4 different size, namely YoloV5s, YoloV5m, YoloV5l, YoloV5x. We combine ExquisiteNetV2 with YoloV5 and test it on Face Mask dataset [30]. Face Mask dataset is originally not divide into training set and test set so we choose image No. 0~49 as test set, the rest is training set.

From Table. , we can learn ExquisiteNetv2-Yolov5 has fastest computing speed and fewest parameters. The ability of detecting objects of ExquisiteNetV2-YoloV5 is better than Yolov5m but the parameters of ExquisiteNetv2-Yolov5 is 20 M less than Yolov5m. The FPS is test on nvidia Geforce 3060, the size of image is 320x320.

Table. 29 result of object detection

|  | mAP 0.5 | Parameter (M) | FPS |
|---|---|---|---|
| Yolov5s | 0.9308 | 7.07 | 172 |
| Yolov5m | 0.9474 | 21.06 |  |
| Yolov5l | 0.9739 | 46.64 |  |
| Yolov5x | 0.9691 | 87.26 |  |
| ExquisiteNetv2-Yolov5 | 0.9434 | **0.95** | **208** |

Fig. 7 result on the test set of Face Mask

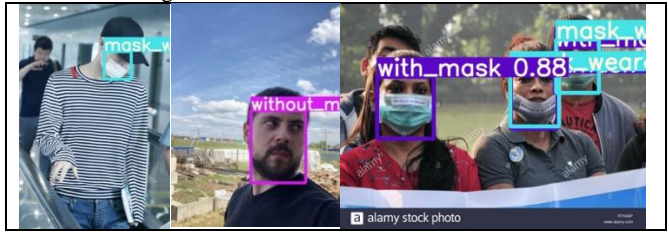

## Conclusion

ExquisiteNetV2 has outstanding ability of analyzing data, fast computing speed, few parameters compared with state-of-the-art models. However, ExquisiteNetV2 is not good at analyzing the very low resolution images. Important of all, according to the experimental results, the performance of ExquisiteNetV2 completely beat ExquisiteNetV1.